\documentclass[sigconf]{acmart}
\usepackage{amsthm,amsmath}
\usepackage{algorithm}
\usepackage{multirow} % Required for multirows
\usepackage{makecell}
\usepackage{soul}
\usepackage{colortbl}
\usepackage{mathtools}
\usepackage{algorithmic}
\usepackage{graphicx}
\usepackage{hyperref}

\AtBeginDocument{%
  \providecommand\BibTeX{{%
    \normalfont B\kern-0.5em{\scshape i\kern-0.25em b}\kern-0.8em\TeX}}}

\copyrightyear{2024}
\acmYear{2024}
\setcopyright{acmlicensed}\acmConference[ICMR '24]{Proceedings of the 2024 International Conference on Multimedia Retrieval}{June 10--14, 2024}{Phuket, Thailand}
\acmBooktitle{Proceedings of the 2024 International Conference on Multimedia Retrieval (ICMR '24), June 10--14, 2024, Phuket, Thailand}
\acmDOI{10.1145/3652583.3658056}
\acmISBN{979-8-4007-0619-6/24/06}

\begin{document}

\title{Learning from Reduced Labels for Long-Tailed Data}

\author{Meng Wei}
\email{weimeng@cumt.edu.cn}
\orcid{0009-0000-3836-6487}
\affiliation{%
  \institution{China University of Mining and Technology}
  \city{Xuzhou}
  \state{Jiangsu}
  \country{China}
  \postcode{221116}
}

\author{Zhongnian Li}
\email{zhongnianli@cumt.edu.cn}
\orcid{0000-0003-3364-8703}
\affiliation{%
	\institution{China University of Mining and Technology}
	\city{Xuzhou}
	\state{Jiangsu}
	\country{China}
	\postcode{221116}
}

\author{Yong Zhou}
\email{yzhou@cumt.edu.cn}
\orcid{0009-0000-3836-6487}
\affiliation{%
	\institution{China University of Mining and Technology}
	\city{Xuzhou}
	\state{Jiangsu}
	\country{China}
	\postcode{221116}
}

\author{Xinzheng Xu}
\authornote{Corresponding author.}
\email{xxzheng@cumt.edu.cn}
\orcid{0000-0001-6973-799X}
\affiliation{%
	\institution{China University of Mining and Technology}
	\city{Xuzhou}
	\state{Jiangsu}
	\country{China}
	\postcode{221116}
}

\renewcommand{\shortauthors}{Meng Wei, Zhongnian Li, Yong Zhou, \& Xinzheng Xu}

\begin{abstract}
  Long-tailed data is prevalent in real-world classification tasks and heavily relies on supervised information, which makes the annotation process exceptionally labor-intensive and time-consuming. Unfortunately, despite being a common approach to mitigate labeling costs, existing weakly supervised learning methods struggle to adequately preserve supervised information for tail samples, resulting in a decline in accuracy for the tail classes. To alleviate this problem, we introduce a novel weakly supervised labeling setting called Reduced Label. The proposed labeling setting not only avoids the decline of supervised information for the tail samples, but also decreases the labeling costs associated with long-tailed data. Additionally, we propose an straightforward and highly efficient unbiased framework with strong theoretical guarantees to learn from these Reduced Labels. Extensive experiments conducted on benchmark datasets including ImageNet validate the effectiveness of our approach, surpassing the performance of state-of-the-art weakly supervised methods. Source code is available at \href{https://github.com/WilsonMqz/LTRL}{https://github.com/WilsonMqz/LTRL}
\end{abstract}





\maketitle

\section{Introduction}
Long-tailed data has received increasing attention in numerous real-world classification tasks \cite{LT_4, LT_5, LT_6, LT_7}. In recent years, deep neural networks have demonstrated remarkable effectiveness across various computer vision tasks. A significant factor contributing to their success has been their inherent scalability \cite{DNN_1, DNN_3, DNN_4}. However, this advantage encounters limitations when these deep neural networks are applied to real-world large-scale datasets, which typically exhibit long-tailed distribution \cite{LT_1, LT_2, LT_3}. In particular, for critical safety- or health-related applications, such as autonomous driving \cite{autonomous_driving} and medical diagnostics \cite{medical_image}, the collected data is inherently characterized by severe imbalance. 

While long-tailed data has received significant attention, the data used requires precise annotation \cite{LT_5, LT_7}. The natural characteristic of long-tailed datasets, which contain a large number of classes, poses a challenge in collecting precise annotations for each sample \cite{LT_5, LT_6, LT_7}. This challenge inevitably leads to massive expenses when considering the utilization of larger datasets, as labeling data usually requires human labor \cite{LTSSL_1}. Hence, decreasing the difficulty of labeling training data has been a long-standing problem in the field of machine learning \cite{SSL_6, pl_2, pl_9}. To alleviate this problem, various weakly supervised learning (WSL) methods have been proposed. Among these methods, semi-supervised learning (SSL) \cite{SSL_1, SSL_2, SSL_3, SSL_5, SSL_6} and partial label learning (PLL) \cite{pl_1, pl_2, pl_4, pl_5, pl_6} are two effective WSL methods for training models without the need for extensive precise annotations. 

Existing SSL methods aim to train effective classifiers with a limited number of labeled samples \cite{LTSSL_5, LTSSL_6}. In contrast, PLL methods alleviate the labeling costs by utilizing less costly partial labels \cite{pl_6, pl_8, pl_9}. Unfortunately, these methods undermine the supervised information for tail classes, which undoubtedly increases the model's challenge in learning the features associated with these classes. Consequently, this motivates us to investigate a novel weakly supervised labeling setting.

\begin{figure*}[!htbp]
	\vspace{-1em}
	\centering
	%\fbox{\rule{0pt}{2in} \rule{.9\linewidth}{0pt}}
	\includegraphics[width=\linewidth]{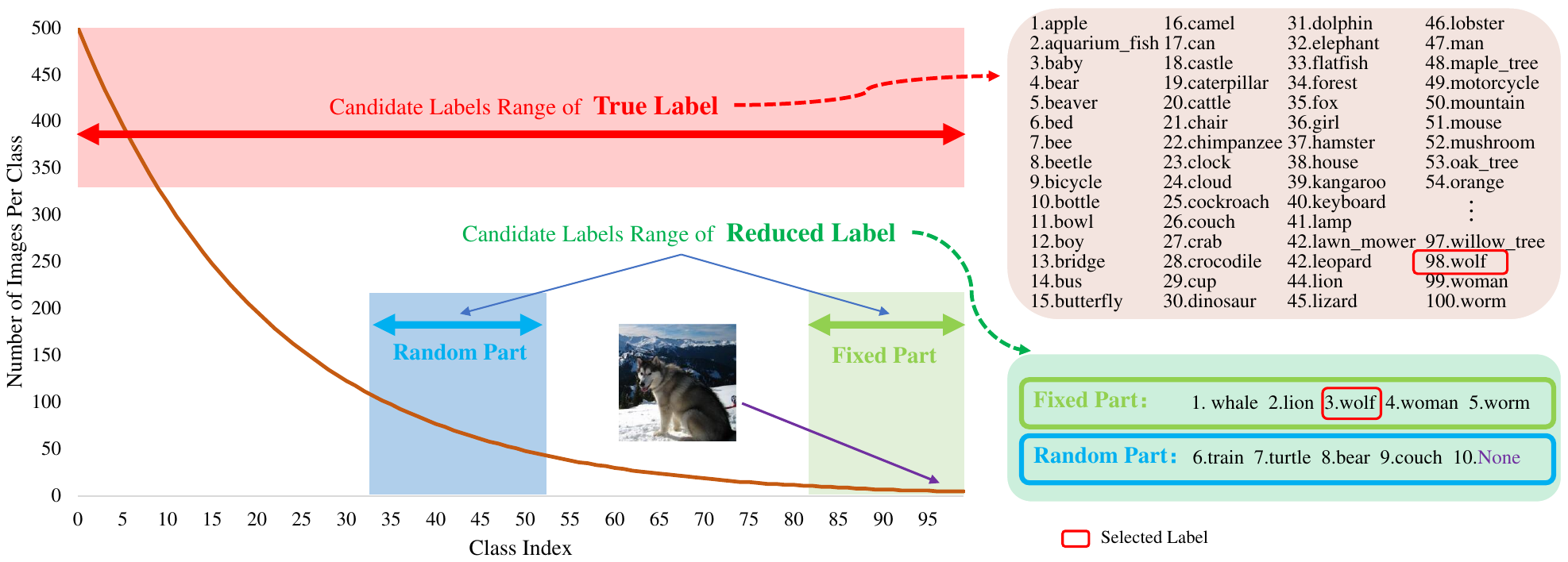}
	\caption{Comparison between \textit{True Label} and \textit{Reduced Label} in CIFAR-100 dataset. Instead of precisely selecting the correct class label from a set of 100 labels, the \textit{Reduced Label} only requires annotators to determine whether the limited set of candidate labels includes the correct class label or not. Here, the correct class label is boxed in red.}
	\label{figure_1}
	\vspace{-1em}
\end{figure*}

In this paper, we consider an alternative weakly supervised labeling setting known as \textit{Reduced Labels (RL)} with less expensive long-tailed data. Instead of selecting all supervised labels from an extensive sequence of classes, RL only requires annotators to determine whether the limited set of candidate labels contains the correct class label or not, as illustrated in Figure \ref{figure_1}. In the context of a large number of classes, the process of selecting the correct class label from an extensive set of candidates is laborious, whereas verifying the presence of the correct label within a smaller set of class labels is more straightforward and thus less costly.

Specifically, as shown in Figure \ref{figure_1}, the RL setting comprises two components: a \textit{fixed part} consisting of tail classes and a \textit{random part} indicating a subset generated by randomly selecting from the set of head classes. Accordingly, instead of precisely selecting correct class label by browsing the entire candidate label set, annotators only need to determine whether the correct class label exists in the reduced labels set with fewer labels or label it as `None' (indicating the absence of the correct class label in reduced labels). Besides, the incorporation of a \textit{fixed part} in reduced labels enables tail samples to acquire correct class labels, ensuring supervised information for tail classes. This constitutes a main advantage distinct from existing weakly supervised approaches, such as PLL methods. For example, as illustrated in Figure \ref{figure_1}, when considering a `wolf' instance within the tail classes, annotators are only required to identify the `wolf' label by examining the reduced labels set that contains a limited number of labels. In summary, this novel setting undoubtedly leads to a substantial reduction in the annotator's time spent reviewing labels, all the while preserving the supervised information of tail class samples. We provide more analysis and description in Section \ref{section_3.5}.

Our contribution in this paper is to provide a straightforward and highly efficient unbiased framework for the classification task with \textit{reduced labels}, denoted as \textit{Long-Tailed Reduced Labels} (LTRL). More specifically, we reformulate the unbiased risk with \textit{reduced labels}, aiming to fully extract and leverage the `None'-labeled data and the limited supervised information from tail classes. Theoretically, we derive the upper bound of the evaluation risk of the proposed method, demonstrating that with an increase in training samples, empirical risk could converge to real classification risk. Extensive experimental results demonstrate that our method exhibits remarkable performance compared to state-of-the-art weakly supervised approaches. Our primary contributions can be summarized as follows:
\begin{enumerate}
	\item [$ \bullet $] A novel labeling setting is introduced, which not only decreases the cost of the labeling task for long-tailed data, but also preserves the supervised information for tail classes.
	\item [$ \bullet $] We design a straightforward and highly efficient unbiased framework tailored for this novel labeling setting and demonstrate that our approach can converge to an optimal state.
	\item [$ \bullet $] Experimental results demonstrate the superiority of the proposed approach over state-of-the-art weakly supervised approaches. 
\end{enumerate}

\section{Related Work}
While the weakly supervised long-tailed scenario has not been explicitly defined in the literature, three closely related tasks are often studied in isolation: semi-supervised learning, long-tailed semi-supervised learning, and partial label learning. Table \ref{table_1} summarizes their differences.

\begin{table*}[!htbp]
	\vspace{-1em}
	\centering
	\renewcommand{\arraystretch}{1}
	\fontsize{8.5}{1em}\selectfont
	\tabcolsep=0.2em
	\caption{Comparison between our method and related existing methods.}
	\begin{tabular}{l|c|c|c|c}%c表示文本居中，c的个数是列数
		\toprule[1pt] 
		\textbf{Task Setting} & \textbf{Lower Labeling Costs} & \textbf{Decrease Browsing Labels} & \textbf{Apply to Long-tailed Data} & \textbf{\makecell{Preserve Tailed Supervised\\Information}}   \\
		\midrule
		\midrule
		Partial Label & $\checkmark$ & $\times$ & $\times$  & $\times$ \\
		\midrule
		Semi-supervised & $\checkmark$ & $\checkmark$ & $\times$ & $\times$   \\
		\midrule
		Long-tailed Semi-supervised &  $\checkmark$ & $\checkmark$ & $\checkmark$ & $\times$ \\
		\midrule
		\textbf{Long-tailed Reduced Label} & $\checkmark$ & $\checkmark$ & $\checkmark$ & $\checkmark$ \\
		\bottomrule[1pt]
	\end{tabular}
	\label{table_1}
	\vspace{-1em}
\end{table*}
\subsection{Semi-supervised Learning}
Semi-supervised learning (SSL) \cite{mixmatch, remixmatch, CR_1, fixmatch, comatch, simmatch, adsh} utilizes labeled and unlabeled data to train an effective classification model, with its major methods are broadly categorized into pseudo-labeling and consistency regularization. Pseudo-labeling methods generate pseudo-labels for unlabeled data and train the model in a supervised manner using these pseudo-labeled instance \cite{LTSSL_1, simmatch}. Consistency regularization methods create diverse augmented versions of unlabeled data and optimize a consistency loss between the unlabeled instances and their augmented counterparts \cite{CR_1, comatch}. Recent approaches combine pseudo-labeling and consistency regularization to enhance model generalization. A notable example is FixMatch \cite{fixmatch}, which outperforms other semi-supervised methods in classification tasks and often serves as a foundational model for generalizing to the long-tailed data.
\begin{figure}[!htbp]
	\centering
	%\fbox{\rule{0pt}{2in} \rule{.9\linewidth}{0pt}}
	
	\includegraphics[width=2.5in]{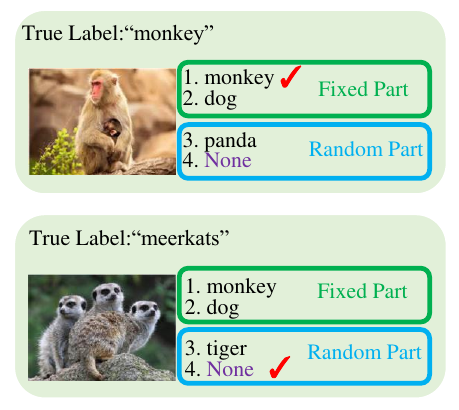}
	\caption{An example of correct class label present or absent in reduced labels set.}
	\label{figure_2}
	\vspace{-1em}
\end{figure}

\begin{figure*}[!htbp]
	\vspace{-1em}
	\centering
	%\fbox{\rule{0pt}{2in} \rule{.9\linewidth}{0pt}}
	\includegraphics[width=.9\linewidth]{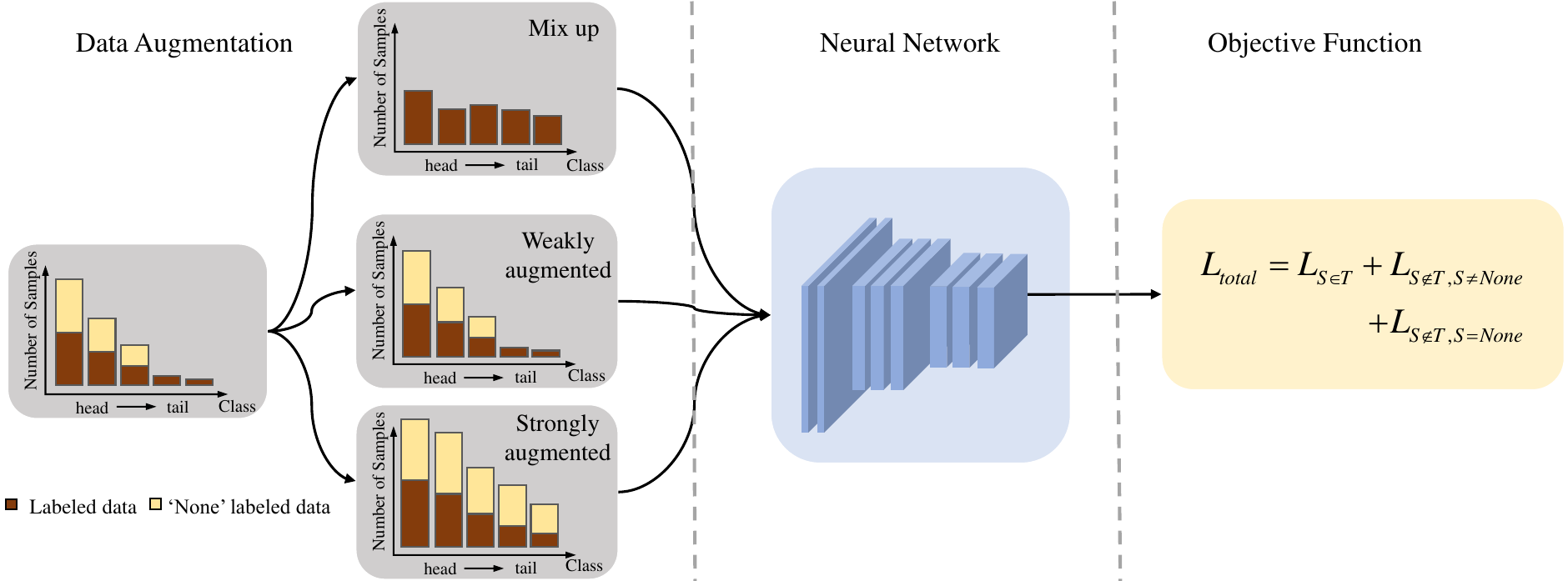}
	\caption{Illustration of the proposed framework.}
	\label{figure_3}
	\vspace{-1em}
\end{figure*}
\subsection{Long-tailed Semi-supervised Learning}
Long-tailed semi-supervised learning (LTSSL) has attracted significant attention in various real-world classification tasks \cite{LTSSL_1, LTSSL_2, LTSSL_3, LTSSL_4, LTSSL_5}. While FixMatch outperforms other SSL methods, adopting a fixed threshold for all classes can lead to excess elimination of correctly pseudo-labeled unlabeled examples in minority classes for class-imbalanced training data. This results in a decrease in recall in minority classes, ultimately diminishing overall performance. Therefore, ADSH \cite{adsh} is proposed to learn from class-imbalanced SSL with adaptive thresholds. However, this method assumes that the distributions of labeled and unlabeled data are almost identical, which is often challenging to maintain in real-world scenarios. Recent work, ACR \cite{LTSSL_6}, presents an adaptive consistency regularizer that effectively utilizes unlabeled data with unknown class distributions, achieving the adaptive refinement of pseudo-labels for various distributions within a unified framework.
\subsection{Partial Label Learning}
Partial label learning (PLL) methods aim to decrease annotation costs \cite{pl_1, pl_3, pl_4, pl_6, pl_8, pl_9}. PLL methods equip each sample with a set of candidate labels, which includes the correct class label. Over the past two decades, numerous practical PLL methods have been proposed. Feng et al. introduce risk-consistency and classifier-consistency models \cite{pl_8}, which accommodate diverse deep networks and stochastic optimizers. In the latest work, Wu et al. revisit the issue of consistent regularization in PLL \cite{pl_9}. They align multiple augmented outputs of an instance to a common label distribution to optimize consistency regularization losses.

In contrast, our approach is applicable to long-tailed data, enhancing the model's performance and generalization ability. Additionally, it provides correct class labels for tail samples, ensuring supervised information for the tail class.

\section{The Proposed Method}
In this section, we introduce an novel labeling setting called \textit{Reduced Label} and present an unbiased algorithm to learn from \textit{reduced labels}.
\subsection{Reduced Label}
In contrast to precisely selecting the correct class label from an exhaustive set of candidate labels, the reduced label involves identifying the correct class label from a limited number of labels or assigning a `None' label if the provided set of reduced labels does not contain the correct class label. As illustrated in Figure \ref{figure_2}, in the case of a `monkey' instance, the annotator is tasked with determining the `monkey' label from the provided reduced labels set. Conversely, when dealing with a `meerkat' instance, if the reduced labels set does not contain the `meerkat' label, the instance is appropriately annotated as `None'. Here, let $\bar{Y}$ denotes the reduced labels set for instance $\textit{\textbf{x}}$ and  $ \bar{Y}  \subseteq \left\{1, 2, \ldots, K\right\} \left(\#\{\bar{Y}\} \ll K\right) $, where $\#\{\bar{Y}\}$ denotes the size of $\bar{Y}$ and $K$ denotes the number of the classes. 

The introduction of reduced labels results in two distinct types of labels: correct class labels and `None' labels. Naturally, two strategies emerge to harmonize the scarcity of labeled data and the weak informativeness of `None' labels: (1) While applying reduced labels to tail samples, priority is given to generating correct class labels;(2) When employing reduced labels on non-tail samples, the choice leans towards producing `None' labels.

Consequently, tail samples with correct class labels help alleviate the scarcity issue, while an abundance of non-tail samples compensates for the weak informativeness of `None' labels. Motivated by these considerations, we fix the tail classes within the provided reduced labels set. As a result, the reduced labels set include fixed part and random part.  Let $\# \{FP\}$ denotes the size of fixed part and $ \# \{RP\}$ denotes the size of random part, i.e. $ \# \{\bar{Y} \} = \left(\# \{FP\} + \# \{RP\} \right) $, where $ FP $ and $RP$ denote the fixed part and random part, respectively.

\subsection{Preliminaries}
Given a $d$-dimensional feature vector $ \textbf{\textit{x}} \in \mathbb{R}^{d} $, and the reduced labels set $ \bar{Y} \subseteq \{1, 2, \ldots, K\} $ associated with instance $ \textbf{\textit{x}} $, where $ \# \{\bar{Y}\} \ll K $ and $ K $ denotes the total number of classes. Let $ S $ indicate the presence of the correct class label within $ \bar{Y} $. Specifically, $ S = j $ signifies that the correct class label $ j $ is present in $ \bar{Y} $, while $ S = None $ indicates the absence of the correct class label, where $ j \in \{1, 2, \ldots, K\} $, and $ S \in \{1, 2, \ldots, K, None\}$. We define the dataset $ \mathcal{D} = \left\{(\textbf{\textit{x}}_i, \bar{Y}_i, S_i) \right\}_{i=1}^{N} $ as being randomly and uniformly sampled from an unknown distribution with density $ P\left(\textbf{\textit{x}}, \bar{Y}, S\right) $. Here, $ N $ is the number of training samples, $ \bar{Y}_i $ denotes the reduced labels set for instance $ \textbf{\textit{x}}_i $, and $ S_i $ denotes the corresponding value of $ S $ for instance $ \textbf{\textit{x}}_i $. Let $ N_k $ denotes the number of samples for class $ k $ in training dataset, we have $ N_1 \geqslant N_2 \geqslant \cdots \geqslant N_K $, and the imbalance radio of dataset is denoted by $ \rho = \frac{N_1}{N_K} $.

To enhance readability, let $ l $ denote the size of the reduced labels set, i.e. $ l = \# \{\bar{Y}\} $. We introduce the multi-class loss function $ \mathcal{L} $ for correct class labels and $ \bar{\mathcal{L}} $ for reduced labels when $ S = None $. Our goal is to learn a multi-class classifier $ f: \textbf{\textit{x}} \mapsto \left\{1, 2, \dots, K \right\} $ that maps instances to class labels, utilizing the provided reduced-labeled samples to minimize the classification risk.

\subsection{Classification from Reduced Labels}
Here, we formulate the problem of reduced labels classification and propose a risk minimization framework. The proposed setting produces both correct class labels and `None' labels. For samples annotated with correct class labels, we derive the following formula:
\begin{small}
	\begin{equation}\label{eq1}
		\begin{aligned}
			P\left(S=j, \textit{\textbf{x}}\right) & = \sum_{i} P\left(S=j, \textit{\textbf{x}} \mid y=i\right) P\left(y=i\right) \\
			& = P
			\left(S=j, \textit{\textbf{x}} \mid y=j\right) P\left(y=j\right) \\
			& = \frac{P\left(y=j \mid S=j, \textit{\textbf{x}}\right) P\left(S=j, \textit{\textbf{x}}\right)}{P\left(y=j\right)} P\left(y=j\right) \\
			& = P(y=j \mid S=j, \textit{\textbf{x}}) P(S=j, \textit{\textbf{x}}).
		\end{aligned}
	\end{equation}
\end{small}
Then, both sides of the equation are simultaneously divided by $ P(S=j, \textit{\textbf{x}}) $, we have
\begin{small}
	\begin{equation}\label{eq2}
		P\left(y=j \mid S = j, \textit{\textbf{x}}\right) = 1,
	\end{equation}
\end{small}
where $ j $ denotes the correct class label. Regarding samples equipped  with `None' labels, let $(K-l)$ denote the size of $j$ for which $j \notin \bar{Y}$. Since the remaining $(K-l)$ labels make an equal contribution to $ P\left(y=j\mid S=None, \textit{\textbf{x}}\right) $, we assume that this subset is drawn independently from an unknown probability distribution with density:
\begin{small}
	\begin{equation}\label{eq3}
		P\left(y=j, \bar{Y} \mid S=None, \textit{\textbf{x}}\right) = \frac{1}{K-l}P\left(\bar{Y} \mid S = None, \textit{\textbf{x}}\right), 
	\end{equation}
\end{small}
where $ K $ and $ l $ denote the size of classes and $ \bar{Y} $, respectively. Fortunately, as illustrated in Section \ref{setion_4}, our experimental results substantiate the reasonableness and validity of this hypothesis.

Subsequently, we derive the following theorem, which models the relationship between correct class label $ j $ and reduced labels set $ \bar{Y} $:
\\
\textbf{Theorem 1.} For any instance $ \textbf{\textit{x}} $ with its correct class label $ y $ and reduced labels set $ \bar{Y} $, the following equality holds:
\begin{small}
	\begin{equation}\label{eq4}
		\begin{aligned}
			P\left(y=j \mid \textbf{\textit{x}}\right) & = P\left(S=j \mid \textit{\textbf{x}}\right) \\
			& + \frac{1}{K-l} \sum_{\bar{Y}} P\left(\bar{Y} \mid S=None, \textbf{\textit{x}}\right)P\left(S=None \mid \textit{\textbf{x}}\right),
		\end{aligned}
	\end{equation}
\end{small}
The proof is presented in Appendix. 

Broadly, our loss can be divided into two components: the tail class loss $\mathcal{L}_{S \in T}$ and the non-tail class loss $\mathcal{L}_{S \notin T}$. The non-tail class loss $\mathcal{L}_{S \notin T}$ originates from a combination of supervised losses characterized by a small number of correct class labels and losses associated with `None' labels. Let us consider the loss of the reduced labels when  $ S = None $. Utilizing Theorem 1, we have the following equation:
\begin{small}
	\begin{equation}\label{eq5}
		\begin{split}
			\bar{\mathcal{L}}\left[f(\textbf{\textit{x}}), \bar{Y}\right] =  \sum_{j \notin \bar{Y}}\mathcal{L}\left[f(\textbf{\textit{x}}), j\right].
		\end{split}
	\end{equation}
\end{small}
This segmentation allows us to effectively manage the impact of diverse classes within our loss framework. Put it together, our total objective function is:
\begin{small}
	\begin{equation}\label{eq6}
		\mathcal{L}_{total} = \mathcal{L}_{S \in T} +  \mathcal{L}_{S \notin T, S \neq None} +  \mathcal{L}_{S \notin T, S = None}.
	\end{equation}
\end{small}
Subsequently, Based on Theorem 1, equation (\ref{eq5}) and equation (\ref{eq6}), an unbiased risk estimator of learning from reduced labels can be derived by the following theorem:
\\
\textbf{Theorem 2.} The classification risk can be described as 
\begin{small}
	\begin{equation}\label{eq7}
		\begin{split}
			R(f) &= \mathbb{E}_{\left(\textit{\textbf{x}}, \bar{Y}, S\right) \sim P\left(\textit{\textbf{x}}, \bar{Y}, S \in T\right)}\mathcal{L}\left[f(\textit{\textbf{x}}), S\right] \\
			& \quad + \mathbb{E}_{\left(\textit{\textbf{x}}, \bar{Y}, S\right) \sim P\left(\textit{\textbf{x}}, \bar{Y}, S \notin T, S \neq None\right)}\mathcal{L}\left[f(\textit{\textbf{x}}), S\right] \\
			& \quad + \mathbb{E}_{\left(\textit{\textbf{x}}, \bar{Y}, S\right) \sim P\left(\textit{\textbf{x}}, \bar{Y}, S \notin T, S = None\right)}\frac{1}{K-l}\bar{\mathcal{L}}\left[f(\textit{\textbf{x}}), \bar{Y}\right].
		\end{split}
	\end{equation}
\end{small}

\subsection{Data Augmentation and Mixup Training}
In order to better utilize the limited information in the reduced labels, we incorporate data augmentation techniques commonly used in SSL methods into our method implementation, including  (1) Random Horizontal Flipping, (2) Random Cropping, and (3) Cutout. These techniques have been shown to achieve surprisingly significant performance in WSL domains \cite{pl_9, fixmatch, LTSSL_6}. Notably, as shown in Figure \ref{figure_3}, for `None'-labeled samples, we employ two data augmentation strategies: a weak augmentation utilizing techniques (1) and (2) and a strong augmentation involving all three aforementioned techniques.

In addition, to further improve the generalization performance of the model, we introduce two mixup methods in our approach: input mixup \cite{input_mixup} and manifold mixup \cite{manifold_mixup}. The introduce of these mixup strategies serves as an effective means of data augmentation, contributing to the regularization of Convolutional Neural Networks (CNNs). Our empirical investigations reveal that the utilization of mixup training yields noteworthy accuracy enhancements, particularly in the context of long-tailed data. To provide a comprehensive understanding of the proposed method, Figure \ref{figure_3} illustrates the framework of the proposed method.
\subsection{Effectiveness Analysis}\label{section_3.5}
Here, we analyze the effectiveness of the reduced labels. Firstly, we investigate its impact on the annotation strength. In order to evaluate the strength of supervised information, we define a metric calculates the proportion of exclusion labels set, defined as $ MS = \frac{L_{dis}}{L_{total}} $, where $ L_{dis} $ denotes the number of discarded options for selecting labels and $ L_{total} $ denotes the number of total options for it. Let $ N_{t} $ and $ N_{nt} $ denotes the size of tail and non-tail samples. Hence, in a dataset with $ K $ classes, the total options $ L_{total} = K \times (N_t + N_{nt})$. For the PLL setting, the discarded options $ L_{dis} = l \times (N_{t} + N_{nt}) $.  But for the proposed setting, the discarded options $ L_{dis} = (l \times N_{nt} + (K-l) \times N_{t}) $, since all tail samples were equipped with correct class label in our setting. Table \ref{table_2} illustrates the results of the strength comparison of the supervised information on the CIFAR-100 dataset, assuming a total number of samples of 20,000, with a tail sample of 5,000. Here, $ \uparrow 5\% $ indicates a $5\%$ improvement in $ MS $ metrics for LTRL method over PLL methods.

Besides, we evaluate the times of browsing labels. We define the metrics $ MB = \frac{C_{RL}}{C_{TL}} $, where $ C_{RL} $ and $ C_{TL} $ denote the times of utilizing reduced labels and all true label. Hence, in a dataset with $ K $ classes, $ C_{TL} = K \times N_{total} $ and $ C_{RL} = l \times N_{total} $, i.e. $ MB = \frac{l}{K} $. Table \ref{table_2} also shows the $ MB $ metrics for the proposed setting compared with fully supervised.
\begin{table}[!htbp]
	\vspace{-0.5em}
	\centering
	\renewcommand{\arraystretch}{0.8}
	\fontsize{8.5}{1em}\selectfont
	\caption{Analysis of effectiveness of reduced labels on CIFAR-100-LT dataset with $\#\{FP\}$ set to 50.}
	\begin{tabular}{l|c|c|c}%c表示文本居中，c的个数是列数
		\toprule[1pt]
		\textbf{Setting} & \textbf{$ MS $ in PLL} & \textbf{$ MS $ in LTRL} & \textbf{$ MB $} \\
		\midrule
		\midrule
		fully supervised & -- & -- & 100$\%$  \\
		$\# \{RP\}$=30 & 80$\%$ & 85$\%$ \ \textcolor{green}{$ \uparrow 5\% $} & 80$\%$  \\
		$\# \{RP\}$=10 & 60$\%$ & 70$\%$ \ \textcolor{green}{$ \uparrow 10\% $} & 60$\%$  \\
		$\# \{RP\}$=5 & 55$\%$ & 66$\%$ \ \textcolor{green}{$ \uparrow 11\% $} & 55$\%$  \\
		\bottomrule[1pt]
	\end{tabular}
	\label{table_2}
	\vspace{-1em}
\end{table}

\begin{table*}[!htbp]
	\vspace{-1em}
	\renewcommand{\arraystretch}{0.8}
	\centering
	\caption{Overview of the five imbalanced datasets used in our experiments.  $ \rho$ indicates the imbalance ratio.}
	\begin{tabular}{l|c|c|c|c|c|c}
		\toprule[1pt]
		\textbf{Dataset} & \textbf{$\#$ Class} & \textbf{$\rho$} & \textbf{Head class size} & \textbf{Tail class size} & \textbf{$ \# $ Training set} & \textbf{$\#$ Test set} \\
		\midrule
		\midrule
		CIFAR-100-LT & 100 & 50 $ \sim $ 100 & 500 & 10 $ \sim $ 5 & 12608 $\sim$ 10847 & 10000 \\
		\midrule
		CIFAR-10-LT & 10 & 50 $ \sim $ 100 & 5000 & 100 $ \sim $ 50 & 13996 $\sim$ 12406 & 10000 \\
		\midrule
		SVHN-LT & 10 & 50 $ \sim $ 100 & 4500 & 90 $ \sim $ 45 & 12596 $\sim$ 11165 & 26032 \\
		\midrule
		STL-10-LT & 10 & 50 $ \sim $ 100 & 500 & 10 $ \sim $ 5 & 1394 $\sim$ 1236 & 8000 \\
		\midrule
		ImageNet-200-LT & 200 & 50 & 500 & 10 & 25082 & 10000 \\
		\bottomrule[1pt]
	\end{tabular}
	\label{table_3}
\end{table*}

\begin{table*}[!htbp]
	\centering
	\renewcommand{\arraystretch}{1}
	\fontsize{9}{1em}\selectfont
	\tabcolsep=0.3em
	\caption{Test accuracy on (a) CIFAR-10-LT, (b) SVHN-LT and (c) STL-10-LT datasets. The best results are in bold.}
	\begin{tabular}{l|cccc|cccc}%c表示文本居中，c的个数是列数
		\toprule[1pt]
		\multirow{3}{*}{\textbf{Method}} & \multicolumn{4}{c|}{\textbf{$\rho$ = 50}} & \multicolumn{4}{c}{\textbf{$\rho$ = 100}} \\
		\cline{2-9}
		~ & $>$ 1000 & $\leqslant$ 1000 $\&$ $>$ 200 & $<$ 200 & & $>$ 1000 & $\leqslant$ 1000 $\&$ $>$ 200 & $<$ 200 & \\
		~ & \textbf{Many-shot} & \textbf{Medium-shot} & \textbf{Few-shot} & \textbf{Overall} & \textbf{Many-shot} & \textbf{Medium-shot} & \textbf{Few-shot} & \textbf{Overall} \\
		
		\midrule
		\midrule
		
		\cellcolor{lightgray!40}Fully Supervised & \cellcolor{lightgray!40}85.70$\pm$0.17  & \cellcolor{lightgray!40}78.50$\pm$0.26 & \cellcolor{lightgray!40}66.45$\pm$0.80 & \cellcolor{lightgray!40}82.05$\pm$0.03 & \cellcolor{lightgray!40}83.86$\pm$0.44  & \cellcolor{lightgray!40}66.83$\pm$0.15 & \cellcolor{lightgray!40}58.25$\pm$1.70 & \cellcolor{lightgray!40}76.91$\pm$0.12 \\
		RC \cite{pl_8} & 77.76$\pm$0.88 & 48.50$\pm$1.22 & --  & 53.43$\pm$0.26 & 72.28$\pm$1.22 & 28.46$\pm$1.86 & 9.55$\pm$0.83 & 47.54$\pm$1.57 \\
		CC \cite{pl_8} & 84.52$\pm$0.39 & 72.46$\pm$0.42 & 58.71$\pm$0.73  & 73.79$\pm$0.49 & 83.41$\pm$0.51 & 65.86$\pm$0.19 & -- & 65.13$\pm$1.59 \\
		PLCR \cite{pl_9} & \textbf{89.44$\pm$1.07} & 39.53$\pm$1.31  & -- & 58.08$\pm$1.17 & \textbf{85.22$\pm$0.53} & 18.88$\pm$0.72  & -- & 49.25$\pm$0.74 \\ 
		FixMatch \cite{fixmatch} & 78.06$\pm$0.89 & 74.87$\pm$1.17 & 62.55$\pm$0.80 & 73.45$\pm$0.69 & 79.52$\pm$1.44 & 66.56$\pm$2.29 & 50.20$\pm$1.40 & 69.47$\pm$0.94  \\
		ADSH \cite{adsh} & 67.68$\pm$1.06 & 81.30$\pm$0.78 & 81.75$\pm$0.75 & 74.11$\pm$2.05 & 71.46$\pm$0.89 & 68.70$\pm$1.78 & 71.00$\pm$0.54 & 71.01$\pm$0.77 \\
		ACR \cite{LTSSL_6} & 84.35$\pm$0.31 & 83.16$\pm$0.26 & 79.15$\pm$0.85 & 82.88$\pm$0.07 & 83.86$\pm$0.14 & 80.36$\pm$0.04 & 69.05$\pm$0.45 & 79.79$\pm$0.04 \\
		\midrule
		LTRL (Ours) & 86.00$\pm$0.32 & \textbf{83.55$\pm$0.75} & \textbf{82.10$\pm$0.40} & \textbf{83.82$\pm$0.38} & 84.52$\pm$0.78 & \textbf{81.14$\pm$1.38} & \textbf{78.07$\pm$1.17} & \textbf{80.39$\pm$0.81} \\
		
		\midrule[1pt]
		
		\multicolumn{9}{c}{ (a) Top-1 classification accuracy on CIFAR-10-LT.} \\
		
		\midrule[1pt]
		
		\multirow{3}{*}{\textbf{Method}} & \multicolumn{4}{c|}{\textbf{$\rho$ = 50}} & \multicolumn{4}{c}{\textbf{$\rho$ = 100}} \\
		\cline{2-9}
		~ & $>$ 1000 & $\leqslant$ 1000 $\&$ $>$ 200 & $<$ 200 & & $>$ 1000 & $\leqslant$ 1000 $\&$ $>$ 200 & $<$ 200 & \\
		~ & \textbf{Many-shot} & \textbf{Medium-shot} & \textbf{Few-shot} & \textbf{Overall} & \textbf{Many-shot} & \textbf{Medium-shot} & \textbf{Few-shot} & \textbf{Overall} \\
		
		\midrule[1pt]
		\midrule[1pt]
		
		\cellcolor{lightgray!40}Fully Supervised & \cellcolor{lightgray!40}94.10$\pm$0.09  & \cellcolor{lightgray!40}86.51$\pm$0.12 & \cellcolor{lightgray!40}71.84$\pm$0.63 & \cellcolor{lightgray!40}91.06$\pm$0.02 & \cellcolor{lightgray!40}92.59$\pm$0.21  & \cellcolor{lightgray!40}82.79$\pm$3.05 & \cellcolor{lightgray!40}70.95$\pm$6.46 & \cellcolor{lightgray!40}89.27$\pm$0.34 \\
		RC \cite{pl_8} & 91.25$\pm$0.78 & 46.66$\pm$0.23 & --  & 69.45$\pm$0.43 & 92.66$\pm$0.32 & 21.26$\pm$1.08 & -- & 64.56$\pm$0.43 \\
		CC \cite{pl_8} & \textbf{96.65$\pm$0.37} & 88.75$\pm$0.43 & 79.52$\pm$2.07  & 90.19$\pm$2.29 & \textbf{96.78$\pm$0.62} & 85.07$\pm$0.84 & -- & 81.64$\pm$0.14\\
		PLCR \cite{pl_9} & 96.20$\pm$0.21 & 30.28$\pm$0.69  & -- & 72.18$\pm$3.17 & 95.93$\pm$0.36 & 30.67$\pm$0.24  & -- & 68.37$\pm$0.31 \\ 
		FixMatch \cite{fixmatch} & 93.83$\pm$0.35 & 89.24$\pm$0.83 & 84.57$\pm$0.63 & 91.37$\pm$0.05 & 93.66$\pm$0.28 & 88.95$\pm$0.92 & 76.80$\pm$2.10 & 90.40$\pm$0.20 \\
		ADSH \cite{adsh} & 92.91$\pm$ 0.21 & 89.55$\pm$0.31 & 84.79$\pm$0.35 & 90.85$\pm$0.13 & 93.24$\pm$0.17 & 89.48$\pm$0.21 & 79.70$\pm$0.30 & 90.62$\pm$0.08 \\
		ACR \cite{LTSSL_6} & 95.01$\pm$0.08 & 90.74$\pm$0.12 & 83.38$\pm$0.19 & 92.60$\pm$0.11 & 94.98$\pm$0.22 & 88.05$\pm$1.35 & 73.23$\pm$0.35 & 91.20$\pm$0.43 \\
		\midrule
		LTRL (Ours) & 95.83$\pm$0.21 & \textbf{91.55$\pm$0.66} & \textbf{86.92$\pm$0.65} & \textbf{93.16$\pm$0.43} & 95.34$\pm$0.41 & \textbf{89.96$\pm$0.61} & \textbf{81.52$\pm$1.60} & \textbf{91.86$\pm$0.11} \\
		
		\midrule[1pt]
		
		\multicolumn{9}{c}{ (b) Top-1 classification accuracy on SVHN-LT.} \\
		
		\midrule[1pt]
		
		\multirow{3}{*}{\textbf{Method}} & \multicolumn{4}{c|}{\textbf{$\rho$ = 50}} & \multicolumn{4}{c}{\textbf{$\rho$ = 100}} \\
		\cline{2-9}
		~ & $>$ 100 & $\leqslant$ 100 $\&$ $>$ 20 & $<$ 20 & & $>$ 100 & $\leqslant$ 100 $\&$ $>$ 20 & $<$ 20 & \\
		~ & \textbf{Many-shot} & \textbf{Medium-shot} & \textbf{Few-shot} & \textbf{Overall} & \textbf{Many-shot} & \textbf{Medium-shot} & \textbf{Few-shot} & \textbf{Overall} \\
		
		\midrule
		\midrule
		
		\cellcolor{lightgray!40}Fully Supervised & \cellcolor{lightgray!40}70.96$\pm$0.33 & \cellcolor{lightgray!40}28.96$\pm$0.84 & \cellcolor{lightgray!40}4.92$\pm$0.11 & \cellcolor{lightgray!40}41.18$\pm$1.53 & \cellcolor{lightgray!40}69.31$\pm$0.76 & \cellcolor{lightgray!40}24.33$\pm$1.39 & \cellcolor{lightgray!40}5.13$\pm$3.07 & \cellcolor{lightgray!40}37.74$\pm$0.23 \\
		RC \cite{pl_8} & 48.18$\pm$0.32 & 23.92$\pm$0.46 &  1.75$\pm$0.10  & 28.04$\pm$0.39 & 48.25$\pm$0.19 & 20.29$\pm$0.27 &  1.50$\pm$0.20 & 27.01$\pm$0.12 \\
		CC \cite{pl_8} & 69.86$\pm$0.53 & 20.92$\pm$0.32 & 0.67$\pm$0.13  & 33.02$\pm$0.16 & 66.37$\pm$0.94 & 15.41$\pm$0.25 & -- & 31.84$\pm$0.39 \\
		PLCR \cite{pl_9} & 63.81$\pm$0.77 & 14.29$\pm$0.18  & -- & 29.08$\pm$0.58 & 69.03$\pm$0.74 & -- & -- & 27.38$\pm$0.12 \\ 
		FixMatch \cite{fixmatch} & 63.91$\pm$0.14 & 20.21$\pm$0.06 & 13.71$\pm$0.13 & 36.20$\pm$0.05 & 59.53$\pm$0.88 & 3.83$\pm$1.46 & 10.62$\pm$0.23 & 29.88$\pm$0.32 \\
		ADSH \cite{adsh} & 65.50$\pm$0.66 & 7.04$\pm$1.13 & 15.54$\pm$0.19 & 36.32$\pm$0.14 & 64.40$\pm$0.42 & 4.25$\pm$0.57 & 12.95$\pm$0.43 & 31.10$\pm$0.38 \\
		ACR \cite{LTSSL_6} & 69.28$\pm$0.52 & 14.58$\pm$1.04 & 15.25$\pm$1.50 & 37.70$\pm$0.44 & 70.31$\pm$0.47 & 4.58$\pm$0.79 & 4.75$\pm$0.13 & 32.39$\pm$0.58 \\
		\midrule
		LTRL (Ours) & \textbf{73.03$\pm$0.22} & \textbf{27.58$\pm$1.24} & \textbf{25.88$\pm$1.87} & \textbf{45.25$\pm$0.72} & \textbf{70.63$\pm$0.42} & \textbf{21.63$\pm$0.37} & \textbf{19.92$\pm$0.47} & \textbf{40.71$\pm$0.29} \\
		
		\bottomrule[1pt]
		
		\multicolumn{9}{c}{ (c) Top-1 classification accuracy on STL-10-LT.} \\

	\end{tabular}
	\label{table_4}
\end{table*}
\begin{table*}[!htbp]
	\vspace{-1em}
	\centering
	\renewcommand{\arraystretch}{1.1}
	\fontsize{9}{1em}\selectfont
	\tabcolsep=0.3em
	\caption{Test accuracy on CIFAR-100-LT dataset. The best results are in bold.}
	\begin{tabular}{ll|cccc|cccc}%c表示文本居中，c的个数是列数
		\toprule[1pt]
		& \multirow{3}{*}{\textbf{Method}} & \multicolumn{4}{c|}{\textbf{$\rho$ = 50}} & \multicolumn{4}{c}{\textbf{$\rho$ = 100}} \\
		\cline{3-10}
		~ & ~ & $>$ 100 & $\leqslant$ 100 $\&$ $>$ 20 & $<$ 20 & & $>$ 100 & $\leqslant$ 100 $\&$ $>$ 20 & $<$ 20 & \\
		~ & ~ & \textbf{Many-shot} & \textbf{Medium-shot} & \textbf{Few-shot} & \textbf{Overall} & \textbf{Many-shot} & \textbf{Medium-shot} & \textbf{Few-shot} & \textbf{Overall} \\
		
		\midrule
		\midrule
		
		\cellcolor{lightgray!40} & \cellcolor{lightgray!40}Fully Supervised & \cellcolor{lightgray!40}72.94$\pm$0.56 & \cellcolor{lightgray!40}46.83$\pm$0.22 & \cellcolor{lightgray!40}21.06$\pm$1.20 & \cellcolor{lightgray!40}51.14$\pm$0.29  & \cellcolor{lightgray!40}73.06$\pm$0.24 & \cellcolor{lightgray!40}42.63$\pm$0.38 & \cellcolor{lightgray!40}11.13$\pm$2.50 & \cellcolor{lightgray!40}45.33$\pm$0.18 \\
		
		\midrule
		
		\multirow{7}{*}{\rotatebox{90}{Case 1}} 
		& RC \cite{pl_8} & 38.94$\pm$0.19 & 9.17$\pm$0.47 & 1.33$\pm$0.63 & 17.48$\pm$0.23 & 35.60$\pm$0.49 & 7.28$\pm$0.39 & 0.43$\pm$0.22 & 15.41$\pm$0.65 \\
		~ & CC \cite{pl_8} & 74.31$\pm$0.15 & 32.88$\pm$0.04 & 0.70$\pm$0.11 & 38.56$\pm$0.31 & 70.51$\pm$0.77 & 20.86$\pm$0.49 & 0.44$\pm$0.12 & 34.89$\pm$0.94 \\
		~ & PLCR \cite{pl_9} & 66.43$\pm$0.51 & 4.83$\pm$0.69 & -- & 25.02$\pm$0.84 & 66.88$\pm$0.72 & -- & -- & 23.41$\pm$0.56 \\ 
		~ & FixMatch \cite{fixmatch} & 68.69$\pm$0.22 & 45.91$\pm$0.56 & 20.03$\pm$0.50 & 45.98$\pm$0.41 & 68.14$\pm$0.17 & 38.97$\pm$0.05 & 10.83$\pm$1.63 & 40.20$\pm$0.21 \\
		~ & ADSH \cite{adsh} & 64.31$\pm$0.73 & 43.60$\pm$1.53 & 19.34$\pm$1.31 & 44.63$\pm$1.24 & 64.14$\pm$0.59 & 37.02$\pm$1.14 & 10.53$\pm$0.46 & 38.74$\pm$1.50 \\
		~ & ACR \cite{LTSSL_6} & 70.94$\pm$0.39 & 49.54$\pm$0.48 & 26.20$\pm$1.50 & 50.59$\pm$0.07 & 69.11$\pm$0.19 & 44.54$\pm$0.59 & 16.83$\pm$0.50 & 44.83$\pm$0.39 \\
		\cline{2-10}
		~ & LTRL (Ours) & \textbf{78.46$\pm$0.33} & \textbf{58.80$\pm$0.84} & \textbf{34.67$\pm$1.53} & \textbf{58.44$\pm$0.18} & \textbf{77.40$\pm$0.25} & \textbf{52.09$\pm$0.10} & \textbf{20.40$\pm$0.56} & \textbf{51.46$\pm$0.31} \\
		
		\midrule
		\midrule
		
		\multirow{7}{*}{\rotatebox{90}{Case 2}} 
		& RC \cite{pl_8} & 20.29$\pm$0.15 & 3.09$\pm$0.04 & 0.67$\pm$0.11 & 8.56$\pm$0.05 & 20.03$\pm$0.77 & 2.37$\pm$0.49 & 0.53$\pm$0.19 & 7.99$\pm$0.20 \\
		~ & CC \cite{pl_8} & 43.08$\pm$0.55 & 10.57$\pm$0.13 & -- & 23.46$\pm$1.97 & 40.87$\pm$0.38 & 5.46$\pm$0.16 & -- & 20.32$\pm$0.27 \\
		~ & PLCR \cite{pl_9} & 37.57$\pm$0.38 & -- & -- & 13.33$\pm$0.61 & 39.20$\pm$0.82 & -- & -- & 13.72$\pm$0.64 \\ 
		~ & FixMatch \cite{fixmatch} & 37.12$\pm$0.53 & 35.42$\pm$1.01 & 28.26$\pm$1.21 & 33.43$\pm$0.39 & 38.34$\pm$0.20 & 30.40$\pm$0.96 & 17.83$\pm$1.19 & 28.51$\pm$0.22 \\
		~ & ADSH \cite{adsh} & 36.14$\pm$0.77 & 37.28$\pm$1.17 & 29.06$\pm$0.65 & 33.65$\pm$0.81 & 38.68$\pm$0.12 & 31.80$\pm$0.08 & 18.13$\pm$0.41 & 29.17$\pm$0.66 \\
		~ & ACR \cite{LTSSL_6} & \textbf{47.97$\pm$0.23} & 43.74$\pm$0.47 & 31.56$\pm$1.13 & 40.31$\pm$0.31 & 48.57$\pm$0.37 & 35.74$\pm$0.61 & 20.96$\pm$1.00 & 34.33$\pm$0.33 \\
		\cline{2-10}
		~ & LTRL (Ours) & 47.20$\pm$0.76 & \textbf{49.54$\pm$0.20} & \textbf{40.27$\pm$0.58} & \textbf{45.70$\pm$0.25} & \textbf{48.86$\pm$0.76} & \textbf{41.71$\pm$0.45} & \textbf{27.50$\pm$1.53} & \textbf{39.95$\pm$0.22} \\
		
		\bottomrule[1pt]
	\end{tabular}
	\label{table_5}
\end{table*}
\subsection{Estimation error bound}
Here, we analyze the generalization estimation error bound for the proposed approach. Let $ f $ be the classification vector function in the hypothesis set $ \mathcal{F} $. Using  $ \varphi_{\mathcal{L}} $ to denote the upper bound of the loss function $ \mathcal{L} $, i.e., $ \mathcal{L}[f(\textbf{\textit{x}}_{i}, S_{i})] \leqslant \varphi_{\mathcal{L}} $, where $ i \in \{1, \ldots, N\} $ and $ S_i \in \{1, \ldots, K\} $. Because $ \varphi_{\mathcal{L}} $ is the upper bound of  loss function $ \mathcal{L} $, the change of $ \mathop{sup}_{f\in\mathcal{F}} \mid \hat{R}_{TL}(f) - R_{TL}(f) \mid $ is no greater than $ \frac{2\varphi_{\mathcal{L}}}{N_{TL}} $ after some $ \textbf{\textit{x}} $ are replaced. Accordingly, using McDiarmid's inequality \cite{mcdiarmid} to $ \mathop{sup}_{f\in\mathcal{F}} \mid \hat{R}_{TL}(f) - R_{TL}(f) \mid $, we have
\begin{small}
	\begin{equation}\label{eq8}
		\begin{split}
			\mathop{sup}_{f\in\mathcal{F}} \mid  \hat{R}_{TL}(f) - R_{TL}(f) \mid  \leqslant  \mathbb{E} \left[ \mathop{sup}_{f\in\mathcal{F}}\left(\hat{R}_{TL}(f) - R_{TL}(f)\right) \right]
			+ \varphi_{\mathcal{L}}\sqrt{\frac{2log\frac{4}{\delta}}{N_{TL}}}.
		\end{split}
	\end{equation}
\end{small}
By symmetrization \cite{rademacher}, we can obtain
\begin{small}
	\begin{equation}\label{eq9}
		\begin{split}
			\mathbb{E} \left[ \mathop{sup}_{f\in\mathcal{F}}\left(\hat{R}_{TL}(f) - R_{TL}(f)\right)\right]
			= 2 \mathfrak{R}_{N_{TL}}\left( \hat{\ell}\circ \mathcal{F} \right)
			\leqslant 2 L_f \mathfrak{R}_{N_{TL}}(\mathcal{F}).
		\end{split}
	\end{equation}
\end{small}

Similarly, the change of $ \mathop{sup}_{f\in\mathcal{F}} \mid \hat{R}_{None}(f) - R_{None}(f) \mid $ is no greater than $ \frac{2\varphi_{\mathcal{L}}}{N_{None}} $ after some $ \textbf{\textit{x}} $ are replaced. Accordingly, using McDiarmid's inequality (McDiarmid, 2013) to $ \mathop{sup}_{f\in\mathcal{F}} \mid \hat{R}_{None}(f) - R_{None}(f) \mid $, we have
\begin{small}
	\begin{equation}\label{eq10}
		\begin{split}
			\mathop{sup}_{f\in\mathcal{F}} \mid  \hat{R}_{None}(f) - R_{None}(f) \mid & \leqslant  \mathbb{E} \left[ \mathop{sup}_{f\in\mathcal{F}}\left(\hat{R}_{None}(f) - R_{None}(f)\right) \right] \\
			& \qquad + \varphi_{\mathcal{L}}\sqrt{\frac{2log\frac{4}{\delta}}{N_{None}}}.
		\end{split}
	\end{equation}
\end{small}
By symmetrization \cite{rademacher}, we can obtain
\begin{small}
	\begin{equation}\label{eq11}
		\begin{split}
			\mathbb{E} \left[ \mathop{sup}_{f\in\mathcal{F}}\left(\hat{R}_{None}(f) - R_{None}(f)\right)\right]
			& = 2 \mathfrak{R}_{N_{None}}\left( \hat{\ell}\circ \mathcal{F} \right) \\
			& \leqslant 2 L_f \mathfrak{R}_{N_{None}}(\mathcal{F}).
		\end{split}
	\end{equation}
\end{small}
By substituting Formula (\ref{eq9}) and (\ref{eq11})  into Formula (\ref{eq8}) and (\ref{eq10}), we can obtain the following lemma.
\\
\textbf{Lemma 3.} For any $ \delta > 0 $, with the probability at least $ 1- \delta / 2 $, we have
\begin{small}
	\begin{equation}\label{eq12}
		\mathop{sup}_{f\in\mathcal{F}} \mid \hat{R}_{TL}(f) - R_{TL}(f) \mid \leqslant 2 L_{f} \mathfrak{R}_{N_{TL}}(\mathcal{F}) + \varphi_{\mathcal{L}}\sqrt{\frac{2log\frac{4}{\delta}}{N_{TL}}},
	\end{equation}
\end{small}
\begin{small}
	\begin{equation}\label{eq13}
		\mathop{sup}_{f\in\mathcal{F}} \mid  \hat{R}_{None}(f) - R_{None}(f) \mid \leqslant 2 L_{f} \mathfrak{R}_{N_{None}}(\mathcal{F}) + \varphi_{\mathcal{L}}\sqrt{\frac{2log\frac{4}{\delta}}{N_{None}}},
	\end{equation}
\end{small}
where  $ \mathfrak{R}_{N_{TL}}(\mathcal{F}) $ and $ \mathfrak{R}_{N_{None}}(\mathcal{F}) $ are the Rademacher complexities \cite{rademacher} of $ \mathcal{F} $ for the sampling of size $ N_{TL} $ from $ P(\textbf{\textit{x}}, \bar{Y}, S \neq None) $, the sampling of size $ N_{None} $ from $ P(\textbf{\textit{x}}, \bar{Y} \mid S = None) $, $ R_{TL}(f) = \mathbb{E}_{\left(\textit{\textbf{x}}, \bar{Y}, S\right) \sim P\left(\textit{\textbf{x}}, \bar{Y}, S \neq None \right)} $, $ R_{None}(f) = \mathbb{E}_{\left(\textit{\textbf{x}}, \bar{Y}, S\right) \sim P\left(\textit{\textbf{x}}, \bar{Y}, S = None \right)} $,  $ \hat{R}_{TL}(f) $ and $ \hat{R}_{None}(f)  $  denotes the empirical risk  of $ R_{TL}(f) $ and $ R_{None}(f) $.

Based on Lemma 3, the estimation error bound can be expressed as follows.
\\
\textbf{Theorem 4.}	For any $ \delta > 0 $, with the probability at least $ 1- \delta / 2 $, we have 
\begin{small}
	\begin{equation}\label{eq14}
		\begin{split}
			R(\hat{f}) - R(f^{\ast})  & \leqslant 4 L_{f} \mathfrak{R}_{N_{TL}}(\mathcal{F}) + 4 L_{f} \mathfrak{R}_{N_{None}}(\mathcal{F})
			\\ 
			& \quad + 2 \varphi_{\mathcal{L}} \sqrt{\frac{2log\frac{4}{\delta}}{N_{TL}}} +  2 \varphi_{\mathcal{L}} \sqrt{\frac{2log\frac{4}{\delta}}{N_{None}}},
		\end{split}
	\end{equation}
\end{small}
where $ \hat{f} $ denotes the trained classifier, $ R(f^{\ast}) = \underset{f\in \mathcal{F}}{min}R(f)  $. The proof is presented in Appendix.

Lemma 3 and Theorem 4 show that our method exists an estimation error bound. With the deep network hypothesis set $ \mathcal{F} $ fixed, we have $ \mathfrak{R}_{N_{TL}}(\mathcal{F}) = \mathcal{O}(1/\sqrt{N_{TL}}) $ and $ \mathfrak{R}_{N_{None}}(\mathcal{F}) = \mathcal{O}(1/\sqrt{N_{None}}) $. Therefore, with  $ N_{TL} \longrightarrow \infty $ and $ N_{None} \longrightarrow \infty $, we have $R(\hat{f}) = R(f^{\ast}) $, which proves that our method could converge to the optimal state. 
\section{Experiments}\label{setion_4}
In this section, we conduct extensive experiments to demonstrate the effectiveness of the proposed approach under various setting.
\subsection{Experimental setup}
\textbf{Datasets.} We employ widely-used benchmark datasets in our experiments, including CIFAR-10-LT \cite{cifar10}, SVHN-LT \cite{svhn}, STL-10-LT \cite{stl-10}, CIFAR-100-LT \cite{cifar100}, and ImageNet-200-LT \cite{imagenet}. The more details of these datasets is provided below.
\begin{enumerate}
	\item [$ \bullet $] CIFAR-100 \citep{cifar100} datasets consisting of 60,000  $ 32 \times 32 \times 3 $ colored images in RGB format. Each image in the dataset is associated with two labels, namely "fine" label and "coarse" label, which represent the fine-grained and coarse-grained categorizations of the image, respectively. Source: \url{http://www.cs.toronto.edu/~kriz/cifar.html}
	\item [$ \bullet $] CIFAR-10 \citep{cifar10}: The CIFAR-10 dataset has 10 classes of various objects: airplane, automobile, bird, cat, etc. This dataset has 50,000 training samples and 10k test samples and each sample is a colored image in $ 32 \times 32 \times 3 $ RGB formats. Source: \url{https://www.cs.toronto.edu/~kriz/cifar.html}
	\item [$ \bullet $] SVHN \citep{svhn} : The SVHN dataset is a street view house number dataset, which is composed of 10 classes. Each sample is a $ 32 \times 32 \times 3 $ RGB image. This dataset has 73,257 training examples and 26,032 test examples. Source: \url{http://ufldl.stanford.edu/housenumbers/}
	\item [$ \bullet $] STL-10 \citep{stl-10} : The images for STL-10 are from ImageNet, with 113,000 RGB images of 96 x 96 resolution, of which 5,000 are in the training set, 8,000 in the test set, and the remaining 100,000 are unlabeled images. Here we only use its training and test sets in our experiments. Source: \url{https://cs.stanford.edu/~acoates/stl10/}
	\item [$ \bullet $] ImageNet-200 \citep{imagenet} : The images in ImageNet-200 are from the ImageNet dataset, which contains 200 categories with 500 training images, 50 validation images, and 50 test images per class, with a size of 64×64 pixels. Source: \url{https://image-net.org/download-images}
\end{enumerate}
Next, we present the experimental settings on these datasets. Table \ref{table_3} provides an overview of the five datasets.
\begin{enumerate}
	\item [$ \bullet $] \textbf{CIFAR-10-LT and SVHN-LT}: We test our method under  $ \# \{FP\} = 4, \# \{RP\} = 1 $. $ \rho $ is set to $ 50 $ and $ 100 $. 
	\item [$ \bullet $] \textbf{STL-10-LT}: Here, we utilize only the labeled data from the STL-10 dataset for training. We test our method under $ \# \{FP\} = 3, \# \{RP\} = 2 $. $ \rho $ is set to $ 50 $ and $ 100 $. 
	\item [$ \bullet $] \textbf{CIFAR-100-LT}: We test our method under the following setting: case 1): $ \# \{FP\} = 50, \# \{RP\} = 30 $; case 2): $ \# \{FP\} = 50, \# \{RP\} = 5 $. $ \rho $ is set to $ 50 $ and $ 100 $. 
	\item [$ \bullet $] \textbf{ImageNet-200-LT}: Following ACR \cite{LTSSL_6}, we down-sample the image size to 32 $ \times $ 32 due to limited resources. We test our method under the following setting: case 1): $ \# \{FP\} = 100, \# \{RP\} = 30 $; case 2): $ \# \{FP\} = 50, \# \{RP\} = 30 $. $ \rho $ is set to $ 50 $. 
\end{enumerate}
\textbf{Compared Methods.} To demonstrate the superiority of our approach, we conducted comparisons with existing SSL approaches, including the classical SSL approach--FixMatch \cite{fixmatch}, the latest imbalanced SSL approach--ADSH \cite{adsh}, the latest LTSSL approach--ACR \cite{LTSSL_6}, as well as existing PLL approaches, including RC \cite{pl_8}, CC \cite{pl_8}, and PLCR \cite{pl_9}. Among all the comparative methods, we employed the same labeling cost as our method. Specifically, for contrastive SSL approaches, we employed the same amount of labeled data. For contrastive PLL approaches, we set the size of partial labels to $ K - l $. Besides, for all comparative methods, including fully supervised methods, we applied consistent data augmentation operation.
\\
\textbf{Implementation.} Following previous work, we implement our method using Wide ResNet-34-10 on all datasets. We use SGD as the optimizer with a momentum of 0.9 and train the network for 200 epochs with a batch size of 64. Following previous work, we evaluate the efficacy of all methods through average top-1 accuracy of overall classes, many-shot classes (class with over 100 training samples), medium-shot classes (class with $20\sim100$ training samples) and few-shot classes (class under $20$ training samples). For every approach, we present both the mean and standard deviation across three distinct and independent trials in our conducted experiments.

\subsection{Result Comparisons}
\textbf{CIFAR-10-LT/SVHN-LT/STL-10-LT.} The comparison results with the latest SSL and PLL methods are displayed in Table \ref{table_4}. The LTRL method consistently outperforms the existing SSL methods and PLL methods. Furthermore, from Table \ref{table_4}, it can be seen that the disadvantages of existing PLL methods, when directly applied to long-tailed data, their predictions inevitably tend to favor the majority class because of the inability to preserve supervised information for tail classes. Here `$-$' indicates that their accuracy is 0.
\begin{table}[!htbp]
	\centering
	\renewcommand{\arraystretch}{1}
	\fontsize{8}{1em}\selectfont
	\tabcolsep=0.2em
	\caption{Test accuracy on ImageNet-200-LT datatset. The abbreviations FS denotes fully supervised.}
	\begin{tabular}{ll|cccc}%c表示文本居中，c的个数是列数
		\toprule[1pt] 
		&\multirow{2}{*}{\textbf{Method}} & $>$ 100 & $\leqslant$ 100 $\&$ $>$ 20 & $<$ 20 & \\
		~ & ~ & \textbf{Many-shot} & \textbf{Medium-shot} & \textbf{Few-shot} & \textbf{Overall} \\
		\midrule
		\midrule
		~ & \cellcolor{lightgray!40}FS & \cellcolor{lightgray!40}52.61$\pm$0.19 & \cellcolor{lightgray!40}28.02$\pm$0.25 & \cellcolor{lightgray!40}15.44$\pm$1.03 & \cellcolor{lightgray!40}35.75$\pm$0.04  \\
		
		\midrule
		
		\multirow{4}{*}{\rotatebox{90}{Case 1} } 
		& FixMatch \cite{fixmatch} & 33.72$\pm$0.14 & 23.81$\pm$0.41 & 15.11$\pm$0.53 & 25.84$\pm$0.66 \\
		~ & ADSH \cite{adsh} & 33.82$\pm$0.45 & 23.61$\pm$0.93 & 14.56$\pm$1.37 & 25.05$\pm$0.95 \\
		~ & ACR \cite{LTSSL_6} & 35.83$\pm$0.12 & 26.56$\pm$0.08 & 16.89$\pm$0.09 & 28.41$\pm$0.08 \\
		\cline{2-6}
		~ & LTRL (Ours) & \textbf{42.54$\pm$0.56} & \textbf{32.68$\pm$0.07} & \textbf{20.39$\pm$1.57} & \textbf{34.51$\pm$0.04} \\
		
		\midrule
		\midrule
		
		\multirow{4}{*}{\rotatebox{90}{Case 2}} 
		& FixMatch \cite{fixmatch} & 29.83$\pm$0.12 & 12.73$\pm$0.09 & 15.27$\pm$1.13 & 21.49$\pm$0.20 \\
		~ & ADSH \cite{adsh} & 29.04$\pm$0.33 & 13.24$\pm$0.58 & 19.55$\pm$0.36 & 21.47$\pm$0.20 \\
		~ & ACR \cite{LTSSL_6} & 34.46$\pm$0.14 & 15.61$\pm$0.19 & 16.33$\pm$0.53 & 23.78$\pm$0.18 \\
		\cline{2-6}
		~ & LTRL (Ours) & \textbf{36.63$\pm$0.27} & \textbf{18.51$\pm$0.63} & \textbf{25.22$\pm$1.61} & \textbf{27.15$\pm$0.49} \\
		
		\bottomrule[1pt]
	\end{tabular}
	\label{table_6}
\end{table}

It is worth noting that both LTRL and ACR achieve better performance than fully supervised methods. Among them, the improvement of LTRL is relatively a bit higher, specifically 1.77$\%$ on CIFAR-10-LT, 2.10$\%$ on SVHN-LT, and 4.07$\%$ on STL-10-LT. We hypothesize that this phenomenon arises from the inherent noise within the dataset itself, and our method provides `None' labels for noisy samples to help the model learn a better generalization on its own after eliminating the incorrect labels.
\\
\textbf{CIFAR-100-LT.} The corresponding results are summarized in Table \ref{table_5}. Remarkably, the LTRL method consistently outperforms existing SSL and PLL methods, particularly demonstrating superior performance in scenarios involving medium-shot classes and few-shot classes. As illustrated in Table \ref{table_5}, the advantages of our method become more pronounced with larger datasets, particularly in enhancing the performance of tail classes. These findings substantiate the superiority of the proposed method.
\\
\textbf{ImageNet-200-LT.} Here, we verify the effectiveness of our approach on a larger dataset. Since the existing PLL methods perform poorly on larger datasets with larger sets of candidate labels, we do not show their effectiveness here. The results in Table \ref{table_6} show that the LTRL method has a greater improvement over all the baseline methods, which validates the effectiveness of our method on larger datasets.
\subsection{Ablation study}
\textbf{Results under various fixed part.} We fix $ \# \{RP\} = 5$ and vary different $ \#\{FP\} $, i.e. case 1): $ \#\{FP\}=50 $; case 2): $ \#\{FP\}=40 $; case 3): $ \#\{FP\}=30 $; case 4): $ \#\{FP\}=20 $; case 5): $ \#\{FP\}=10 $. Table \ref{table_7} shows the results for different settings. As the supervised information is further decreasing, the accuracy decreases. 
\begin{table}[!htbp]
	\centering
	\renewcommand{\arraystretch}{0.8}
	\fontsize{9}{1em}\selectfont
	\tabcolsep=0.3em
	\caption{Test accuracy on CIFAR-100-LT dataset with different size of fixed part.}
	\begin{tabular}{c|cccc}
		\toprule[1pt]
		\multirow{3}{*}{\textbf{Setting}} & \multicolumn{4}{c}{\textbf{$\rho$ = 50}}  \\
		\cline{2-5} 
		~ & $>$ 100 & $\leqslant$ 100 $\&$ $>$ 20 & $<$ 20 & \\
		~ & \textbf{Many-shot} & \textbf{Medium-shot} & \textbf{Few-shot} & \textbf{Overall} \\
		
		\midrule
		\midrule
		
		Case 1 & 45.61$\pm$0.37 & 49.04$\pm$0.30 & 38.83$\pm$0.52 & 44.27$\pm$0.02 \\
		Case 2 & 45.08$\pm$0.24 & 30.95$\pm$0.39 & 42.15$\pm$0.55 & 38.91$\pm$0.56 \\
		Case 3 & 38.20$\pm$0.56 & 15.12$\pm$0.87 & 45.00$\pm$2.00 & 31.71$\pm$0.28 \\
		Case 4 & 34.26$\pm$0.31 & 11.57$\pm$0.60 & 27.27$\pm$0.39 & 24.99$\pm$0.76 \\
		Case 5 & 22.43$\pm$0.72 & 1.09$\pm$0.21 & 7.60$\pm$1.53 & 11.80$\pm$0.76 \\
		
		\bottomrule
	\end{tabular}
	\label{table_7}
\end{table}
\\
\textbf{Training with/without mixup.} We test several possible values of the hyper-parameter $ \alpha $ for the Beta distribution in mixup techniques. The results of the mixup method experiments are shown in Table \ref{table_8}. It can be observed from Table \ref{table_8} that input mixup (IM) yields better results compared to manifold mixup (MM) and the baseline.
\begin{table}[!htbp]
	\centering
	\renewcommand{\arraystretch}{0.8}
	\fontsize{9}{1em}\selectfont
	\tabcolsep=0.1em
	\caption{Test accuracy on CIFAR-100-LT dataset with/without mixup. $ \alpha $ is the hyperparameter of the Beta distribution in mixup. The abbreviations IM and MM denote input mixup and manifold mixup, respectively.}
	\begin{tabular}{l|cccc}%c表示文本居中，c的个数是列数
		\toprule[1pt] 
		\multirow{3}{*}{\textbf{Setting}} & \multicolumn{4}{c}{\textbf{$\rho$ = 50}}  \\
		\cline{2-5} 
		~ & $>$ 100 & $\leqslant$ 100 $\&$ $>$ 20 & $<$ 20 & \\
		~ & \textbf{Many-shot} & \textbf{Medium-shot} & \textbf{Few-shot} & \textbf{Overall} \\
		
		\midrule
		\midrule
		
		Non-mixup  & 44.94$\pm$0.76 & 45.44$\pm$0.20 & 37.63$\pm$0.58 & 42.17$\pm$0.25 \\
		
		\midrule
		
		IM ($\alpha$=1) & 45.03$\pm$0.25 & 48.23$\pm$0.38 & 37.90$\pm$0.50 & 44.01$\pm$0.02 \\
		IM ($\alpha$=2) & \textbf{45.61$\pm$0.37} & \textbf{49.04$\pm$0.30} & \textbf{38.83$\pm$0.52} & \textbf{44.27$\pm$0.02} \\
		
		\midrule
		
		MM ($\alpha$=1)  & 42.16$\pm$0.36 & 45.46$\pm$0.21 & 35.23$\pm$0.74 & 42.40$\pm$0.33 \\
		MM ($\alpha$=2)  & 40.71$\pm$0.72 & 45.06$\pm$1.16 & 37.10$\pm$2.00 & 40.04$\pm$0.56 \\
		
		\bottomrule[1pt]
	\end{tabular}
	\label{table_8}
\end{table}
\begin{table}[!htbp]
	\vspace{-1em}
	\centering
	\renewcommand{\arraystretch}{0.8}
	\fontsize{9}{1em}\selectfont
	\tabcolsep=0.1em
	\caption{Test accuracy of training with/without fine-tuning after mixup on CIFAR-100-LT dataset.}
	\begin{tabular}{l|cccc}
		\toprule[1pt] 
		\multirow{3}{*}{\textbf{Setting}} & \multicolumn{4}{c}{\textbf{$\rho$ = 50}}  \\
		\cline{2-5} 
		~ & $>$ 100 & $\leqslant$ 100 $\&$ $>$ 20 & $<$ 20 & \\
		~ & \textbf{Many-shot} & \textbf{Medium-shot} & \textbf{Few-shot} & \textbf{Overall} \\
		
		\midrule
		\midrule
		IM ($\alpha$=1) & 45.03$\pm$0.25 & 48.23$\pm$0.38 & 37.90$\pm$0.50 & 44.01$\pm$0.02 \\
		ft+IM ($\alpha$=1) & \textbf{45.60$\pm$0.32} & \textbf{49.00$\pm$0.40} & \textbf{38.17$\pm$0.34} &\textbf{44.40$\pm$0.47} \\
		
		\midrule
		
		IM ($\alpha$=2) & 45.61$\pm$0.37 & 49.04$\pm$0.30 & 38.83$\pm$0.52 & 44.27$\pm$0.02 \\
		ft+IM ($\alpha$=2)  & \textbf{47.20$\pm$0.76} & \textbf{49.54$\pm$0.20} & \textbf{40.27$\pm$0.58} & \textbf{45.70$\pm$0.25} \\
		
		\midrule
		MM ($\alpha$=1)  & 42.16$\pm$0.36 & 45.46$\pm$0.21 & 35.23$\pm$0.74 & 42.40$\pm$0.33 \\
		ft+MM ($\alpha$=1)  & \textbf{42.34$\pm$0.18} & \textbf{45.79$\pm$0.20} & \textbf{35.63$\pm$0.27} & \textbf{41.84$\pm$0.05} \\
		
		\midrule
		
		MM ($\alpha$=2)  & 40.71$\pm$0.72 & 45.06$\pm$1.16 & 37.10$\pm$2.00 & 40.04$\pm$0.56 \\
		ft+MM ($\alpha$=2)  & \textbf{41.94$\pm$0.33} &  \textbf{47.77$\pm$0.06} &  \textbf{37.50$\pm$0.20} & \textbf{41.20$\pm$0.03} \\
		
		\bottomrule[1pt]
	\end{tabular}
	\label{table_9}
	\vspace{-1em}
\end{table}
\\
\textbf{Fine-tuning after mixup.} \cite{fine-tuning} showed that the results of models trained using mixup can be enhanced further by removing mixup in the final few epochs. In our experiments, we initially train with mixup, and then fine-tuning the mixup-trained model for a few epochs for further improvement, which is called ``ft + IM/MM". The results of fine-tuning after mixup training are shown in Table \ref{table_9}. It is evident that both input mixup and manifold mixup, when used in post-training fine-tuning, lead to additional enhancements.
\section{Conclusion}
To alleviate the labeling cost for long-tailed data, we introduce a novel WSL labeling setting. The proposed setting not only alleviates the difficulty of labeling long-tailed data but also preserves the supervised information for tail classes. Furthermore, we propose an unbiased framework with strong theoretical guarantees and validate its effectiveness through extensive experiments on benchmark datasets. In summation, the proposed approach stands as a promising resolution to decreasing the cost of labeling long-tailed data and contributing to the advancement of weakly supervised learning and labeling techniques.
\section{Acknowledgement}
This work was supported by the National Natural Science Foundation of China (No. 61976217, 62306320), the Natural Science Foundation of Jiangsu Province (No. BK20231063), the Graduate Innovation Program of China University of Mining and Technology (No. 2024WLKXJ188), the Postgraduate Research $\&$ Practice Innovation Program of Jiangsu Province.
\bibliographystyle{ACM-Reference-Format}
\bibliography{sample-base}

\appendix

\section{Proof of Theorem 1}
\textit{\textbf{Proof.}}
In the proposed setting, let $ S $ indicate the presence of the correct class label within $ \bar{Y} $. Specifically, $ S = j $ indicates that the correct class label $ j $ is present in $ \bar{Y} $, while $ S = None $ indicates the absence of the correct class label, where $ j \in \{1, 2, \ldots, K\} $. Let us models the relationship between correct class label $ j $ and reduced labels set $ \bar{Y} $, we have
\begin{equation}\label{eq_2}
	\begin{aligned}
		P\left(y=j \mid \textbf{\textit{x}}\right) & =  \sum_{k=1}^{K} P\left(y=j, S=k \mid \textbf{\textit{x}}\right) + P\left(y=j, S= None \mid \textbf{\textit{x}}\right) \\
		& = \sum_{k \neq j}^{K} P\left(y=j, S=k \mid  \textbf{\textit{x}}\right) \\
		& +  P\left(y=j, S= j \mid \textbf{\textit{x}}\right) + P\left(y=j, S= None \mid \textbf{\textit{x}}\right) \\
		& = P\left(y=j, S=j \mid \textbf{\textit{x}}\right) +  P\left(y=j, S= None \mid \textbf{\textit{x}}\right) \\
		& \qquad  \left(\because \sum_{k \neq j} P\left(y=j, S=k \mid \textit{\textbf{x}}\right) = 0\right) \\
		& = P\left(y=j \mid S = j, \textbf{\textit{x}}\right)P\left(S=j \mid \textit{\textbf{x}}\right) \\
		& + P\left(y=j\mid S= None, \textbf{\textit{x}} \right)P\left(S=None \mid \textit{\textbf{x}}\right).
	\end{aligned}
\end{equation}

Here, $ S = None $ denotes the reduced labels set $ \bar{Y} $ do not have correct class label. For the right half of the above equation, we denote  $ A =  P\left(y=j \mid S= None, \textbf{\textit{x}}\right)P\left(S=None \mid \textit{\textbf{x}}\right) $.
Since all $ P\left(\bar{Y} \mid S=None, \textit{\textbf{x}}\right) $ for $ j \notin \bar{Y} $ make an equal contribution to $ P\left(y=j, \bar{Y} \mid S=None, \textit{\textbf{x}}\right) $, we assume that this subset is drawn independently from an unknown probability distribution with density:
\begin{equation}\label{eq_3}
	P\left(y=j, \bar{Y} \mid S=None, \textit{\textbf{x}}\right) = \frac{1}{K-l}P\left(\bar{Y} \mid S = None, \textit{\textbf{x}}\right), 
\end{equation}
where $ K $ and $ l $ denote the size of classes and $ \bar{Y} $, respectively. 

Then, we have
\begin{equation}\label{eq_4}
	\begin{aligned}
		A & =  P\left(y=j \mid S= None, \textbf{\textit{x}}\right)P\left(S=None \mid \textit{\textbf{x}}\right) \\
		& = \sum_{\bar{Y}} P\left(y=j, \bar{Y} \mid S=None, \textbf{\textit{x}}\right)P\left(S=None \mid \textit{\textbf{x}}\right) \\
		& = \frac{1}{K-l} \sum_{\bar{Y}} P\left(\bar{Y} \mid S=None, \textbf{\textit{x}}\right)P\left(S=None \mid \textit{\textbf{x}}\right).
	\end{aligned}
\end{equation}

By substituting Formula (\ref{eq_4}) and Formula (\ref{eq_3}) into Formula (\ref{eq_2}), we can obtain
\begin{equation}
	\begin{aligned}
		P\left(y=j \mid \textbf{\textit{x}}\right) & = P\left(S=j \mid \textit{\textbf{x}}\right) \\
		& + \frac{1}{K-l} \sum_{\bar{Y}} P\left(\bar{Y} \mid S=None, \textbf{\textit{x}}\right)P\left(S=None \mid \textit{\textbf{x}}\right),
	\end{aligned}
	\nonumber
\end{equation}
which concludes the proof of Theorem 1. \hfill $\square$

\section{Proof of Theorem 2}
\textit{\textbf{Proof.}}
Let $ M = P(\textit{\textbf{x}}) $, according to the Theorem 1, we have
\begin{equation}\label{eq_5}
	\begin{aligned}
		R(f) & =  \mathbb{E}_{\textbf{\textit{x}} \sim M} \sum_{j}^{K}P(j \mid \textbf{\textit{x}}) \mathcal{L}[f(\textbf{\textit{x}}), j] \\
		& = \mathbb{E}_{\textbf{\textit{x}} \sim M}\left\{\sum_{j \in T}P(j \mid \textbf{\textit{x}})\mathcal{L}[f(\textbf{\textit{x}}), j] + \sum_{j \notin T}P(j \mid \textbf{\textit{x}})\mathcal{L}[f(\textbf{\textit{x}}), j] \right\} \\
		& = \mathbb{E}_{(\textit{\textbf{x}}, \bar{Y}, S) \sim P\left(\textit{\textbf{x}}, \bar{Y}, S \in T\right)}\mathcal{L}\left[f(\textit{\textbf{x}}), S\right] + \mathbb{E}_{\textbf{\textit{x}} \sim M}\sum_{j \notin T}P(j \mid \textbf{\textit{x}})\mathcal{L}[f(\textbf{\textit{x}}), j].
	\end{aligned}
\end{equation}

For the right part of the above equation, we denote $ B = \qquad $ $ \mathbb{E}_{\textbf{\textit{x}} \sim M}\sum_{j \notin T}P(j \mid \textbf{\textit{x}})\mathcal{L}[f(\textbf{\textit{x}}), j] $. Then we have
\begin{equation}\label{eq_6}
	\begin{aligned}
		B & = \mathbb{E}_{\textit{\textbf{x}} \sim M}\sum_{j \notin T} \mathcal{L}\left[f(\textit{\textbf{x}}), j\right] \\
		& \qquad \left\{ \sum_{k}^{K} \left[P(y=j, S=k \mid \textit{\textbf{x}})\right] + P(y=j, S=None \mid \textit{\textbf{x}}) \right\}  \\
		& = \mathbb{E}_{\textit{\textbf{x}} \sim M}\sum_{j \notin T} \left\{ \sum_{k \neq j}^{K} \left[P(y=j, S=k \mid \textit{\textbf{x}})\right] + P(y=j, S=j \mid \textit{\textbf{x}}) \right.  \\
		& \qquad + P(y=j, S=None \mid \textit{\textbf{x}}) \Biggr\} \mathcal{L}\left[f(\textit{\textbf{x}}), j\right]  \\
		& = \mathbb{E}_{\textit{\textbf{x}} \sim M}\sum_{j \notin T} \left[P(y=j, S=j \mid \textit{\textbf{x}}) \right. \\
		&\qquad + P(y=j, S=None \mid \textit{\textbf{x}})] \mathcal{L}\left[f(\textit{\textbf{x}}), j\right] \\
		& \qquad \left(\because \sum_{k \neq j}^{K} \left[P(y=j, S=k \mid \textit{\textbf{x}})\right] = 0\right)  \\
		& = \mathbb{E}_{\textit{\textbf{x}} \sim M}\sum_{j \notin T}P(y=j \mid S=j, \textit{\textbf{x}})P(S=j \mid \textit{\textbf{x}}) \mathcal{L}\left[f(\textit{\textbf{x}}), j\right] \\
		& \qquad + \mathbb{E}_{\textit{\textbf{x}} \sim M}\sum_{j \notin T}P(y=j \mid S=None, \textit{\textbf{x}})P(S=None \mid \textit{\textbf{x}}) \mathcal{L}\left[f(\textit{\textbf{x}}), j\right] \\
		& = \mathbb{E}_{\textit{\textbf{x}} \sim M} \sum_{j \notin T} P(S=j \mid \textit{\textbf{x}}) \mathcal{L}\left[f(\textit{\textbf{x}}), j\right]  \left(\because P\left(y=j \mid S = j, \textit{\textbf{x}}\right) = 1 \right)\\
		& \qquad + \mathbb{E}_{\textit{\textbf{x}} \sim M}\sum_{j \notin T}P(y=j \mid S=None, \textit{\textbf{x}})P(S=None \mid \textit{\textbf{x}}) \mathcal{L}\left[f(\textit{\textbf{x}}), j\right] \\
		& = \mathbb{E}_{(\textit{\textbf{x}}, \bar{Y}, S) \sim P(\textit{\textbf{x}}, \bar{Y}, S \notin T, S \neq None)} \mathcal{L}\left[f(\textit{\textbf{x}}), S\right]  \\
		& \qquad+ \mathbb{E}_{\textit{\textbf{x}} \sim M}\sum_{j \notin T}P(y=j \mid S=None, \textit{\textbf{x}})P(S=None \mid \textit{\textbf{x}}) \mathcal{L}\left[f(\textit{\textbf{x}}), j\right]. \\
	\end{aligned}
\end{equation}

For the right part of the above equation, we denote $ C =\qquad $ $ \mathbb{E}_{\textit{\textbf{x}} \sim M}\sum_{j \notin T}P(y=j \mid S=None, \textit{\textbf{x}})P(S=None \mid \textit{\textbf{x}}) \mathcal{L}\left[f(\textit{\textbf{x}}), j\right] $. Then, we can obtain
\begin{equation}\label{eq_7}
	\begin{aligned}
		C & = \mathbb{E}_{\textit{\textbf{x}} \sim M}\sum_{j \notin T}P(y=j\mid S=None, \textit{\textbf{x}})P(S=None \mid \textit{\textbf{x}}) \mathcal{L}\left[f(\textit{\textbf{x}}), j\right] \\
		& =  \mathbb{E}_{\textit{\textbf{x}} \sim M}\sum_{j \notin T} \sum_{\bar{Y}} P(y=j, \bar{Y} \mid S=None, \textit{\textbf{x}})P(S=None \mid \textit{\textbf{x}}) \mathcal{L}\left[f(\textit{\textbf{x}}), j\right] \\
		& = \mathbb{E}_{\textit{\textbf{x}} \sim M} \\
		& \sum_{\bar{Y}} \left\{\sum_{j \notin T,j \in \bar{Y}}P(y=j, \bar{Y} \mid S=None, \textit{\textbf{x}}) P(S=None \mid \textit{\textbf{x}}) \mathcal{L}\left[f(\textit{\textbf{x}}), j\right] \right. \\
		& + \left. \sum_{j \notin T, j \notin \bar{Y}} P(y=j, \bar{Y} \mid S=None, \textit{\textbf{x}}) P(S=None \mid \textit{\textbf{x}}) \mathcal{L}\left[f(\textit{\textbf{x}}), j\right] \right\}. \\
	\end{aligned}
\end{equation}

Since  $ S = None $ denotes the reduced labels set $ \bar{Y} $ do not have correct class label $ j $, $ j $ and `$None$' cannot coexist. Accordingly, we can obtain $ \sum_{j \in \bar{Y}} P(y=j, \bar{Y} \mid S=None, \textit{\textbf{x}}) = 0 $. Consequently, for the left part of the above equation, we have  $ \sum_{j \notin T, j \in \bar{Y}} P(y=j, \bar{Y} \mid S=None, \textit{\textbf{x}}) P(S=None \mid \textit{\textbf{x}}) \mathcal{L}\left[f(\textit{\textbf{x}}), j\right] = 0 $. Besides, there is a fixed part (Tail classes) in $\bar{Y}$, i.e. $ T \subseteq \bar{Y}$. if $j \notin \bar{Y}$, then $ j \notin T$, so we have $\sum_{j \notin T, j \in \bar{Y}} = \sum_{ j \notin \bar{Y}}$. Then, we can obtain
\begin{equation}\label{eq_8}
	\begin{aligned}
		C & = \mathbb{E}_{\textit{\textbf{x}} \sim M} \sum_{\bar{Y}} \sum_{j \notin T, j \notin \bar{Y}} P(y=j, \bar{Y} \mid S=None, \textit{\textbf{x}})P(S=None \mid \textit{\textbf{x}}) \mathcal{L}\left[f(\textit{\textbf{x}}), j\right] \\
		& = \mathbb{E}_{\textit{\textbf{x}} \sim M} \sum_{\bar{Y}} \sum_{j \notin \bar{Y}}  \frac{1}{K-l} P(\bar{Y} \mid S=None, \textit{\textbf{x}}) P(S=None \mid \textit{\textbf{x}}) \mathcal{L}\left[f(\textit{\textbf{x}}), j\right] \\
		& = \mathbb{E}_{\textit{\textbf{x}} \sim M} \frac{1}{K-l} \sum_{\bar{Y}} P(\bar{Y} \mid S=None, \textit{\textbf{x}}) P(S=None\mid \textit{\textbf{x}})\sum_{j \notin \bar{Y}} \mathcal{L}\left[f(\textit{\textbf{x}}), j\right] \\
		& = \mathbb{E}_{\textit{\textbf{x}} \sim M} \frac{1}{K-l} \sum_{\bar{Y}} P(\bar{Y} \mid S=None, \textit{\textbf{x}}) P(S=None\mid \textit{\textbf{x}}) \bar{\mathcal{L}}\left[f(\textit{\textbf{x}}), \bar{Y}\right] \\
		& \qquad \qquad \left(\bar{\mathcal{L}}\left[f(\textit{\textbf{x}}), \bar{Y}\right] = \sum_{j \notin \bar{Y}} \mathcal{L}\left[f(\textit{\textbf{x}}), j\right] \right) \\
		& = \mathbb{E}_{(\textit{\textbf{x}}, \bar{Y}, S) \sim P(\textit{\textbf{x}}, \bar{Y}, S \notin T, S=None)} \frac{1}{K-l}\bar{\mathcal{L}} \left[f(\textit{\textbf{x}}), \bar{Y}\right].
	\end{aligned}
\end{equation}

By substituting Formula (\ref{eq_8})  into Formula (\ref{eq_6}), we have 
\begin{equation}\label{eq_9}
	\begin{aligned}
		B & = \mathbb{E}_{(\textit{\textbf{x}}, \bar{Y}, S) \sim P(\textit{\textbf{x}}, \bar{Y}, S \notin T, S \neq None)} \mathcal{L}\left[f(\textit{\textbf{x}}), S\right] \\
		& \qquad +  \mathbb{E}_{(\textit{\textbf{x}}, \bar{Y}, S) \sim P(\textit{\textbf{x}}, \bar{Y}, S \notin T, S=None)} \frac{1}{K-l}\bar{\mathcal{L}} \left[f(\textit{\textbf{x}}), \bar{Y}\right]
	\end{aligned}
\end{equation}

By substituting Formula (\ref{eq_9})  into Formula (\ref{eq_5}), we can obtain 
\begin{equation}
	\begin{aligned}
		R(f) & = \mathbb{E}_{(\textit{\textbf{x}}, \bar{Y}, S) \sim P\left(\textit{\textbf{x}}, \bar{Y}, S \in T\right)}\mathcal{L}\left[f(\textit{\textbf{x}}), S\right] \\
		& \qquad + \mathbb{E}_{(\textit{\textbf{x}}, \bar{Y}, S) \sim P(\textit{\textbf{x}}, \bar{Y}, S \notin T, S \neq None)} \mathcal{L}\left[f(\textit{\textbf{x}}), S\right] \\
		& \qquad +  \mathbb{E}_{(\textit{\textbf{x}}, \bar{Y}, S) \sim P(\textit{\textbf{x}}, \bar{Y}, S \notin T, S=None)} \frac{1}{K-l}\bar{\mathcal{L}} \left[f(\textit{\textbf{x}}), \bar{Y}\right]
	\end{aligned}
	\nonumber
\end{equation}
which proves the Theorem 2. \hfill $\square$

\section{Proof of Theorem 4}
\textit{\textbf{Proof.}}
Let us divide $ R(\hat{f}) - R(f^*) $ into two parts, then we have
\begin{equation}\label{eq_16}
	\begin{split}
		R(\hat{f}) - R(f^*) & = \left(R_{TL}(\hat{f}) + R_{None}(\hat{f}) \right) - \Big( R_{TL}(f^*) + R_{None}(f^*) \Big) \\
		& = \left( R_{TL}(\hat{f}) - R_{TL}(f^*) \right)+ \left( R_{None}(\hat{f}) - R_{None}(f^*) \right)
	\end{split}
\end{equation}

It is intuitive to obtain
\begin{equation}\label{eq_17}
	\begin{split}
		& R_{TL}(\hat{f}) - R_{TL}(f^*) \\
		& = \hat{R}_{TL}(\hat{f}) - \hat{R}_{TL}(f^*) + R_{TL}(\hat{f}) - \hat{R}_{TL}(\hat{f}) + \hat{R}_{TL}(f^*)-R_{TL}(f^*) \\ 
		& \leqslant R_{TL}(\hat{f}) - \hat{R}_{TL}(\hat{f}) + \hat{R}_{TL}(f^*)-R_{TL}(f^*) \\
		& \qquad  \left(\because \hat{R}_{TL}(\hat{f}) - \hat{R}_{TL}(f^*) \leqslant 0\right)  \\
		& \leqslant 2\mathop{sup}_{f\in\mathcal{F}} \mid  \hat{R}_{TL}(f) - R_{TL}(f) \mid \\
		& \leqslant 4L_f \mathfrak{R}_{N_{TL}}(\mathcal{F})
		+ 2\varphi_{\mathcal{L}}\sqrt{\frac{2log\frac{4}{\delta}}{N_{TL}}}.
	\end{split}
\end{equation}

Similarly, 
\begin{equation}\label{eq_18}
	\begin{split}
		& R_{None}(\hat{f}) - R_{None}(f^*)\\
		& = \hat{R}_{None}(\hat{f}) - \hat{R}_{None}(f^*) + R_{None}(\hat{f}) - \hat{R}_{None}(\hat{f}) \\
		& \qquad + \hat{R}_{None}(f^*)-R_{None}(f^*) \\ 
		& \leqslant R_{None}(\hat{f}) - \hat{R}_{None}(\hat{f}) + \hat{R}_{None}(f^*)-R_{None}(f^*) \\
		& \leqslant 2\mathop{sup}_{f\in\mathcal{F}} \mid  \hat{R}_{None}(f) - R_{None}(f) \mid \\
		& \leqslant 4 L_f \mathfrak{R}_{N_{None}}(\mathcal{F})
		+ 2\varphi_{\mathcal{L}}\sqrt{\frac{2log\frac{4}{\delta}}{N_{None}}}.
	\end{split}
\end{equation}

By substituting Formula (\ref{eq_18}) and Formula (\ref{eq_17})  into Formula (\ref{eq_16}), we can obtain  
\begin{equation}
	\begin{split}
		R(\hat{f}) - R(f^*) & = \left( R_{TL}(\hat{f}) - R_{TL}(f^*) \right)+ \left( R_{None}(\hat{f}) - R_{None}(f^*) \right) \\
		& \leqslant 2\mathop{sup}_{f\in\mathcal{F}} \mid  \hat{R}_{TL}(f) - R_{TL}(f) \mid \\ 
		& \qquad + 2\mathop{sup}_{f\in\mathcal{F}} \mid  \hat{R}_{None}(f) - R_{None}(f) \mid \\
		& \leqslant  4 L_f \mathfrak{R}_{N_{TL}}(\mathcal{F}) + 4L_f \mathfrak{R}_{N_{None}}(\mathcal{F}) \\
		& \qquad + 2\varphi_{\mathcal{L}}\sqrt{\frac{2log\frac{4}{\delta}}{N_{TL}}} + 2\varphi_{\mathcal{L}}\sqrt{\frac{2log\frac{4}{\delta}}{N_{None}}},
	\end{split}
	\nonumber
\end{equation}
which concludes the proof of Theorem 4. \hfill $\square$

\end{document}